\documentclass[conference]{IEEEtran}
\IEEEoverridecommandlockouts

\usepackage{cite}
\usepackage{amsmath,amssymb,amsfonts}
\usepackage{graphicx}
\usepackage{textcomp}
\usepackage[table]{xcolor}
\usepackage{booktabs}
\usepackage{array}
\usepackage{multirow}
\usepackage{url}
\graphicspath{{figures/}}

\definecolor{tablehead}{HTML}{EEF2F7}
\definecolor{tablesubtle}{HTML}{F8FAFC}

\def\BibTeX{{\rm B\kern-.05em{\sc i\kern-.025em b}\kern-.08em
    T\kern-.1667em\lower.7ex\hbox{E}\kern-.125emX}}

\newcommand{\method}{CAVE-Mem}
\newcommand{\score}{\mathrm{score}}
\newcommand{\valid}{\mathrm{valid}}

\begin{document}

\title{CAVE-Mem: Boundary-Aware Experience Validation for Memory Search}

\author{\IEEEauthorblockN{Xinyu Li}
\IEEEauthorblockA{Kent State University}}

\maketitle

\begin{abstract}
Long-term memory agents increasingly rely on iterative search and reusable
experience to answer questions over large personal, factual, or narrative
histories. However, current experience-memory systems largely optimize
\emph{relevance}: they retrieve past search lessons that appear similar to the
current state and inject them into the prompt. A relevant experience can still
be harmful when the memory substrate, question intent, answer granularity, or
evidence boundary changes. We propose \method{}, a training-free framework that
represents experience as a typed intervention operator with applicability,
boundary, and utility conditions. \method{} first obtains a base memory-search
answer, then allows an operator to change it only if the operator matches the
current substrate, answer contract, evidence boundary, and cross-fitted
utility; otherwise the system abstains. Experiments across long-term
conversational memory, multi-hop question answering, and long-document
narrative reasoning show consistent gains over relevance-only experience reuse.
\end{abstract}

\begin{IEEEkeywords}
agent memory, memory search, experience reuse, retrieval augmented generation
\end{IEEEkeywords}

\section{Introduction}

Long-term memory is a core capability of agentic systems. A memory agent must
preserve fine-grained historical information while answering questions that may
require temporal reasoning, multi-hop entity tracking, exact slot extraction,
or reasoning over very long contexts. This need connects classic external
memory models~\cite{graves2012long,graves2014neural,weston2014memory,
sukhbaatar2015end}, modern retrieval-augmented generation
systems~\cite{zhu2020retrieval,guu2020retrieval,karpukhin2020dense,
izacard2021leveraging}, and recent agent-memory architectures for persistent
interaction~\cite{packer2023memgpt,park2023generative,chhikara2025mem0,
xu2026mem,kang2025memory}. Recent systems further show that memory agents
can search over historical context iteratively and can reuse procedural
experience from previous trajectories~\cite{yan2025general,wang2026r}.

However, experience reuse introduces a validity problem that is not captured
by retrieval relevance alone. A retrieved experience may be semantically
similar to the current question and still be inappropriate for the current
memory substrate, answer contract, or evidence boundary. Prior work on
long-context evaluation and parametric/non-parametric memory already shows
that more context or more retrieved text is not automatically easier for a
model to use~\cite{mallen2023not,liu2023lost,bai2023longbench,
yen2025helmet}. Our experiments show that the same issue appears inside
experience reuse. A temporal operator can help when the memory substrate
contains state transitions but mislead when the substrate is narrative prose; a
counting operator can help when evidence enumerates instances but hurt when the
question asks for duration or frequency; and an answer-contract operator can
help direct slot questions while dropping essential modifiers in causal
questions. These errors are not just retrieval failures. They are
\emph{validity-boundary failures}: the system stores what worked without
storing where it is allowed to work.

\begin{figure}[t]
\centering
\includegraphics[width=\columnwidth]{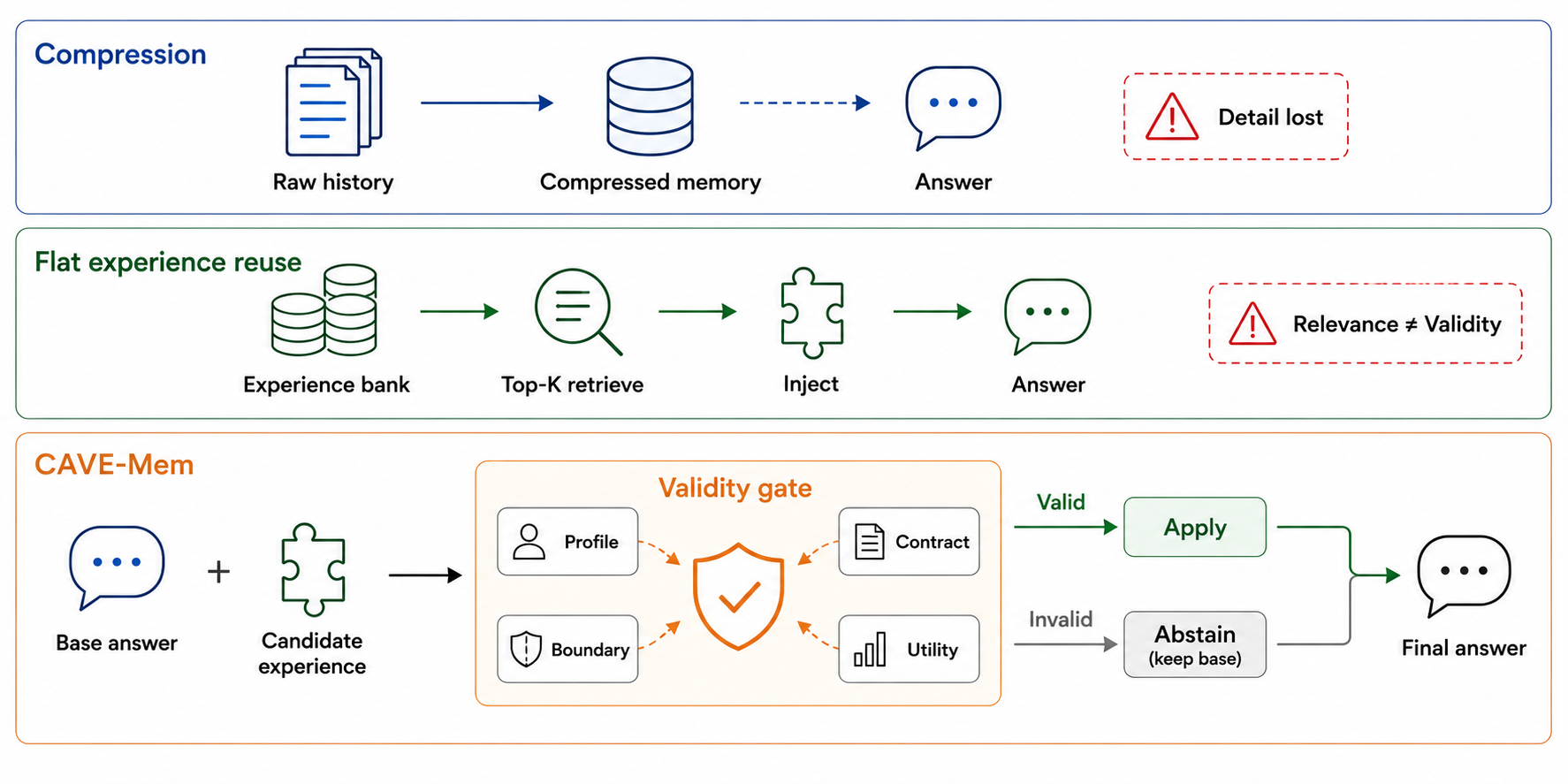}
\caption{From memory compression to flat experience reuse and validated
intervention. \method{} validates a candidate experience before it can change
the base answer; otherwise, it abstains and preserves the base answer.}
\label{fig:paradigm}
\end{figure}

We propose \method{} (Conditional Applicability and Validated Experience for
Memory Search), a training-free gate for experience reuse. \method{} wraps a
base memory-search agent, profiles the memory substrate, infers the answer
contract induced by the question, and instantiates typed experience operators.
An operator can affect the answer only when it is compatible with the
substrate, preserves the requested answer contract, stays inside the evidence
boundary established by the search trajectory, and has positive held-out
utility under matched diagnostics. Otherwise, the system abstains to the base
answer. Concrete operators such as temporal-state resolution, exact-slot
restoration, or answer canonicalization instantiate these checks; they are not
conversation-specific output patches.

We evaluate \method{} on matched long-memory, multi-hop QA, and narrative QA
settings. Across LoCoMo, HotpotQA, and NarrativeQA, \method{} improves over
retrieval, deep-search, and experience-reuse baselines. The largest gains occur
on LoCoMo, where temporal and exact-slot failures are frequent.

This paper makes three contributions:
\begin{itemize}
    \item \textbf{Problem formulation.} We formulate experience reuse as a
    conditional intervention decision rather than a pure retrieval problem.
    \item \textbf{Method.} We introduce \method{}, which represents retrieved
    experience as typed operators governed by substrate, contract, boundary,
    utility, and abstention checks.
    \item \textbf{Evaluation.} We report matched evaluations across long-term
    conversational memory, multi-hop question answering, and narrative
    reasoning, with ablations that isolate the contribution of each check.
\end{itemize}

\section{Related Work}

\subsection{Iterative Deep Memory Search for LLM Agents}
Deep memory search systems avoid compressing the entire history into a single
static representation. Instead, they preserve raw or lightly processed
historical contexts and perform iterative search at inference time. This
setting builds on retrieval-augmented and open-domain QA systems that combine
sparse retrieval, dense retrieval, and generative readers~\cite{
robertson2009probabilistic,zhu2020retrieval,guu2020retrieval,
karpukhin2020dense,izacard2021leveraging,reimers2019sentence,
chen2024bge}. Prior memory-search systems instantiate this paradigm with
iterative planning, retrieval, evidence integration, reflection, and reusable
trajectory-level experience~\cite{yan2025general,wang2026r}. These systems motivate a broader
design question for memory agents: once experience is available, what makes it
valid for the current search state?

\subsection{Experience Learning and Agent Self-Evolution}
Agent self-improvement methods externalize reflections, skills, or procedural
knowledge from past interactions. These methods show that agents can benefit
from previous trajectories without end-to-end policy training, including
reasoning-action traces, verbal reinforcement, self-feedback, tool-use traces,
and skill libraries~\cite{shinn2024reflexion,
madaan2023self,wang2023voyageropenendedembodiedagent}.
Generative-agent and memory-operating-system work further shows that persistent
memory can support long-horizon personalization and simulation
behavior~\cite{park2023generative,packer2023memgpt,li2025memos}. Existing
experience memories mainly emphasize how to externalize, retrieve, or organize
reusable guidance. \method{} studies a distinct control problem: whether a
retrieved guidance item should be allowed to intervene under the current
question, evidence, and answer contract.

\subsection{Validity, Negative Transfer, and Process Critique Signals}
Experience can be harmful when transferred outside its validity domain. For
memory search, validity depends on the question type, the memory substrate, the
available evidence, and the expected answer granularity. Rubric-based
diagnosis, reflective critique, and counterfactual replay provide useful
signals for identifying beneficial behavior, but a memory agent still needs an
inference-time mechanism for suppressing invalid guidance. Our work focuses on
this mechanism: explicit applicability predicates, boundary vetoes, utility
diagnostics, and abstention. This framing is also aligned with recent memory
systems that organize evolving user or environment histories, but our focus is
not memory storage itself; it is whether a retrieved experience is valid before
intervention~\cite{chhikara2025mem0,xu2026mem,rasmussen2025zep,
kang2025memory,yan2026memory,banerjee2026apex}.

\section{Preliminary}

\subsection{Deep Memory Search}
We model memory search as an iterative process over a memory store $M$. Given
a question $q$, the agent repeatedly performs planning, searching, integration,
and reflection:
\begin{align}
    p_t &= f_{\mathrm{plan}}(q, z_t),\\
    r_t &= f_{\mathrm{search}}(p_t, M),\\
    z_{t+1} &= f_{\mathrm{integrate}}(q, z_t, r_t),\\
    c_t &= f_{\mathrm{reflect}}(q, z_{t+1}).
\end{align}
Here $z_t$ is the temporary working memory and $c_t$ decides whether the agent
should stop or continue searching.

\begin{figure}[t]
\centering
\includegraphics[width=\columnwidth]{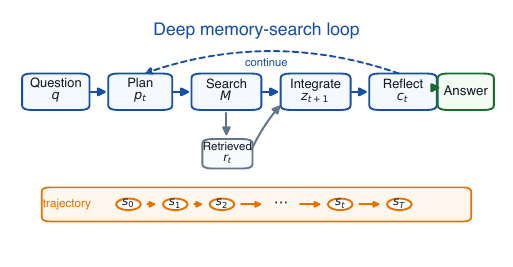}
\caption{Deep memory search iterates over planning, memory search, integration,
and reflection, producing retrieved evidence $r_t$ and a state trajectory
$s_0,\ldots,s_T$.}
\label{fig:deepsearch}
\end{figure}

\subsection{Memory Search Trajectory}
A memory search trajectory records not only the final answer, but also the
sequence of intermediate decisions that produced it:
\begin{equation}
\tau = (s_0, s_1, \ldots, s_T), \quad
s_t=(q_t, a_t, z_t, r_t),
\end{equation}
where $q_t$ is the current request, $a_t$ is the planning, searching, or
reflection action, $z_t$ is the temporary memory, and $r_t$ is the retrieved
evidence. Such trajectories provide the substrate from which procedural
experience can be distilled, retrieved, and reused. \method{} focuses on the
intervention decision: if an experience is retrieved for a new state $s_t$,
should it be allowed to change the agent's answer?

\subsection{Experience as Intervention}
We use \emph{experience} to denote operator-level procedural memory distilled
from prior search trajectories. An experience may be a natural-language search
lesson, a transformation over a working-memory state, or a bounded evidence
restoration operator; in all cases, it is an action that can intervene in a
new trajectory rather than merely a text snippet appended to a prompt. Let
$E=\{e_i\}$ be an experience bank. Standard retrieval selects
\begin{equation}
e_{1:K} = \operatorname{TopK}_{e\in E} \score(e, q, z_t)
\end{equation}
and injects the selected experience into the prompt. This optimizes relevance
but not validity. We instead define candidate experience as a typed
intervention whose utility is
\begin{equation}
\Delta(e;q,M)=R(A(q,M,e))-R(A(q,M,\varnothing)),
\end{equation}
where $A$ is the base agent and $R$ is the task metric. Since direct
counterfactual evaluation is expensive during inference, \method{} estimates
whether $\Delta$ is positive using applicability predicates, boundary checks,
and cached utility diagnostics.

\section{Methodology}

\subsection{Overview of the Framework}
\begin{figure*}[t]
\centering
\includegraphics[width=0.96\textwidth]{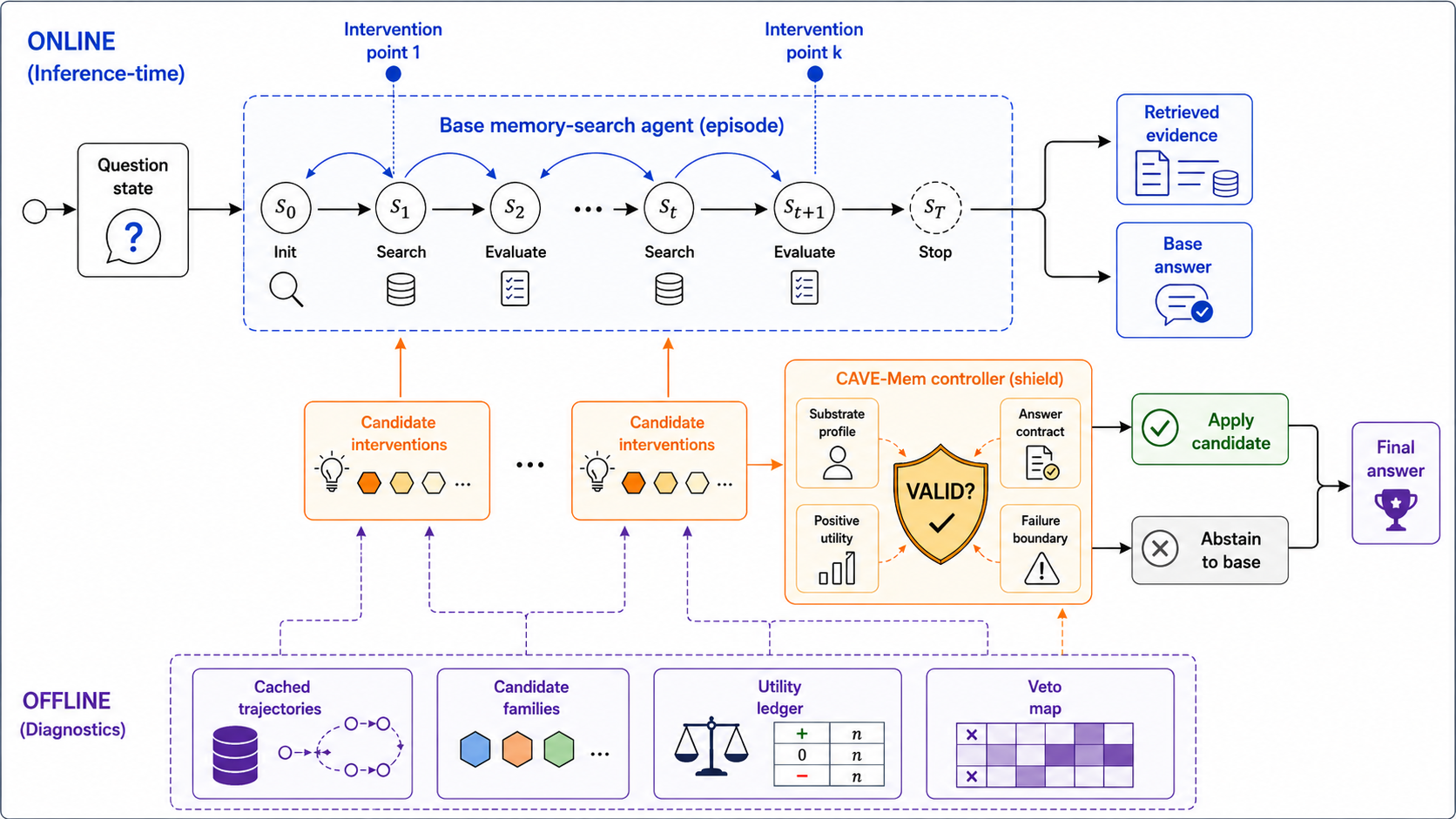}
\caption{\method{} framework. Offline diagnostics produce cached trajectories,
candidate families, a utility ledger, and a veto map. Online, the base
memory-search agent produces retrieved evidence and a base answer, while
candidate interventions are routed through the \method{} validity-gate
controller/shield. The gate checks Profile, Contract, Boundary, and Utility
before choosing between applying the candidate and abstaining to the base
answer.}
\label{fig:framework}
\end{figure*}

\method{} is an inference-time validity gate placed on top of a deep
memory-search agent. In Fig.~\ref{fig:framework}, the same gate is drawn as the
\method{} controller/shield. It observes the question, the base answer, the
retrieved evidence, and a set of typed experience interventions. The gate then
asks whether each intervention is valid under the current Profile, Contract,
Boundary, and Utility checks.
Formally, \method{} augments each candidate
experience intervention with five fields:
\begin{equation}
e=(g,\rho,\kappa,b,\hat{u}),
\end{equation}
where $g$ is the operator body, such as natural-language guidance, a
working-memory transformation, or a bounded answer transformation,
$\rho$ is a substrate-profile predicate, $\kappa$ is an answer-contract
predicate, $b$ is a failure-boundary predicate, and $\hat{u}$ is observed
utility from diagnostics. These four predicates correspond to the gate checks
shown in Fig.~\ref{fig:framework}. The intervention is valid only if
\begin{equation}
\begin{aligned}
\valid(e,q,z,m)=&
\mathbb{1}[\rho(q,z,m)=1]\cdot \mathbb{1}[\kappa(q,z,g)=1]\\
&\cdot \mathbb{1}[b(q,z,m,g)=0]\cdot \mathbb{1}[\hat{u}>0].
\end{aligned}
\end{equation}
Here $z$ denotes the working-memory and retrieved-evidence state produced by the
base agent, and $m$ is the memory substrate profile. If no intervention is valid,
the system abstains to the base answer.

\subsection{Principled Gate Design}
\method{} is built from validity invariants of memory search. An experience
operator should be reused only when four conditions hold: the memory substrate
can support the operator, the operator preserves the answer contract requested
by the question, the operator stays within the evidence boundary created by the
search trajectory, and the operator has positive held-out utility under matched
diagnostics. These conditions are independent of a particular benchmark label
or conversation identifier. The concrete operator families used in our
implementation are surface realizations of the invariants: temporal-state
operators preserve state transitions, slot-restoration operators preserve
answer contracts, quote/reason operators preserve answer-bearing spans, and
utility operators suppress high-variance interventions. The method therefore
learns when an experience is allowed to intervene; it is not a universal output
editor that rewrites every answer.

\subsection{Substrate Profile}
The same experience operator can have different effects depending on the memory
substrate. \method{} profiles the memory as episodic dialogue, fact-pack QA, or
narrative text. Date and count anchors are reused on episodic dialogue and
fact-pack memory when their conditions hold, but are suppressed on narrative
memory unless the answer contract is sufficiently narrow. This prevents
episodic temporal operators from being transferred unconditionally to narrative
reasoning.

\subsection{Answer Contract}
The question defines an expected answer shape. Direct slot questions, such as
``who'', ``where'', ``when'', and ``how many'', often reward concise answers.
Explanation questions require reason or event phrases and are more vulnerable
to over-compression. \method{} therefore accepts a narrower operator only when
it is textually anchored in the base answer, agreed on by independent candidate
paths, and consistent with the question type.

\subsection{Failure Boundary}
Candidate interventions are rejected when they violate evidence or contract
boundaries. Examples include converting a full date to a bare year when the
question asks ``when'', introducing a year not supported by the question or
base answer, dropping essential modifiers, selecting an off-slot attribute,
losing temporal relations, or replacing a reason/quote/feeling with an
unsupported paraphrase. These checks instantiate the general requirement that
an intervention must not change the semantic object being asked about. The
validity gate is intentionally conservative: uncertain candidates are not
applied.

\subsection{Positive Utility}
The utility ledger stores the observed effect of each operator family under
matched diagnostics, but it is used in a cross-fitted manner. For a target
conversation or evaluation block $d$, the cached utility $\hat{u}_{-d}$ is
estimated only from diagnostic trajectories outside $d$. An operator family is
allowed to pass the gate only when this held-out cached effect is positive
under the current substrate and answer contract. The target example's gold
answer, correction outcome, and conversation-level aggregate score are never
queried by the gate that decides whether to change that example. This is the
source of the utility predicate in Fig.~\ref{fig:framework}: a candidate can be
relevant and satisfy local boundary checks, but still be blocked if
cross-fitted diagnostics show that the corresponding family is not beneficial.

\subsection{Deep Search with Validated Experience}
\method{} does not assume a single monolithic experience item. During online
deep search, the base memory-search agent first produces a search trajectory,
retrieved evidence, and a base answer. The validity gate then instantiates
experience operators from four schema-level sources. First, the base answer
provides a conservative fallback that preserves the full search trajectory.
Second, answer-contract operators canonicalize the base answer only when the
question asks for a narrow slot. Third, evidence-boundary operators inspect the
cited pages for exact dates, counts, reasons, feelings, and quoted phrases that
are often lost during summary integration. Fourth, cross-fitted utility
diagnostics record which operator families have positive empirical effect
under the current memory substrate without using the target block being
answered.

The final intervention is selected by a validity-bounded objective:
\begin{equation}
e^\star=\arg\max_{e\in C(q,z,M)}
\valid(e,q,z,m)\cdot\left(\score(e,q,z)+\hat{u}_{e,-d}\right),
\end{equation}
where $C(q,z,M)$ is the candidate set. If all candidates have zero validity,
the system abstains to the base answer. The final policy is therefore a
composition of candidate generation and validity-gated selection, rather than a
larger prompt that asks the model to ``try harder''.

\begin{table*}[t]
\caption{Gate checks used by \method{} to convert retrieved experience into
validated interventions.}
\label{tab:predicates}
\centering
\footnotesize
\setlength{\tabcolsep}{4.8pt}
\renewcommand{\arraystretch}{1.16}
\begin{tabular*}{0.98\textwidth}{@{\extracolsep{\fill}}
>{\raggedright\arraybackslash}p{0.145\textwidth}
>{\raggedright\arraybackslash}p{0.285\textwidth}
>{\raggedright\arraybackslash}p{0.285\textwidth}
>{\raggedright\arraybackslash}p{0.17\textwidth}}
\toprule
\rowcolor{tablehead}
\textbf{Gate check} & \textbf{Valid intervention} & \textbf{Boundary veto} &
\textbf{Role in \method{}} \\
\midrule
\rowcolor{tablesubtle}
\textbf{Substrate profile} & Operator family is enabled for the current memory
substrate, e.g., episodic dialogue, fact-pack QA, or narrative text. &
Operator transfers a substrate-specific procedure outside its profiled domain.
& Prevents cross-substrate transfer. \\
\textbf{Answer contract} & Candidate directly fills the requested slot, e.g., person,
place, date, count, or object. & Candidate answers a nearby attribute or
compresses a reason into an unsupported phrase. & Aligns answer shape with the
question. \\
\rowcolor{tablesubtle}
\textbf{Temporal boundary} & Candidate preserves the exact date or state transition
anchored in retrieved evidence. & Candidate replaces a constrained date with a
broader month/year or an unsupported current state. & Prevents temporal drift. \\
\textbf{Count boundary} & Candidate is backed by enumerated instances with stable
wording. & Candidate uses vague forms such as ``at least'' when an exact count
is required. & Converts evidence into exact counts. \\
\rowcolor{tablesubtle}
\textbf{Quote/reason boundary} & Candidate preserves source-grounded reason, feeling,
or quoted wording. & Candidate paraphrases away the answer-bearing phrase. &
Protects short answer-bearing spans. \\
\textbf{Utility ledger} & Operator family has positive cross-fitted diagnostic
effect under the current substrate. & Candidate source is high-variance under
matched counterfactual replay. & Selects high-utility interventions. \\
\bottomrule
\end{tabular*}
\end{table*}

\subsection{Abstention as a First-Class Action}
A key design choice is to make abstention a normal outcome of experience
reuse. If a candidate is relevant but its conditions are not satisfied,
\method{} abstains to the base memory-search answer. This differs from flat
experience retrieval, where retrieved text necessarily changes the prompt and
can alter the search plan even when the experience is only superficially
similar. In our implementation, abstention occurs at several points: before
candidate generation when the substrate profile is incompatible, after
candidate generation when the answer contract is broad, and after scoring when
the candidate belongs to a family with negative measured utility.

\section{Experiments}

\subsection{Research Questions}
We evaluate \method{} through three research questions.
\begin{itemize}
    \item \textbf{RQ1: Overall Performance.} Does validated experience reuse
    outperform strong retrieval, deep-search, and experience-reuse baselines?
    \item \textbf{RQ2: Scaling Behavior.} Does the method remain effective
    across smaller and stronger Qwen backbones?
    \item \textbf{RQ3: Ablation Study.} Which validation components explain
    the gains, and how does abstention prevent negative transfer?
\end{itemize}

\subsection{Datasets and Metrics}
We evaluate on LoCoMo,
HotpotQA~\cite{yang2018hotpotqa}, and
NarrativeQA~\cite{kovcisky2018narrativeqa}. These datasets cover complementary
stressors: long-term conversational memory, multi-hop Wikipedia question
answering, and long-document narrative reading. They are also representative
of broader long-context and memory evaluation trends, including LongBench,
LongMemEval, Lost-in-the-Middle, HELMET, and LoCoMo-Plus~\cite{
bai2023longbench,wu2024longmemeval,liu2023lost,yen2025helmet,
li2026locomo}. LoCoMo uses nine held-out conversations
(\texttt{conv-30/41/42/43/44/47/48/49/50}); \texttt{conv-26} is excluded
because it is used as the LoCoMo demonstration source. HotpotQA uses the
128-question subsets under \texttt{eval\_400}, \texttt{eval\_1600}, and
\texttt{eval\_3200}, corresponding to the 56K, 224K, and 448K context
regimes. NarrativeQA uses a seed-42 300-question subset with average context
length about 87K tokens. LoCoMo is
evaluated with token F1 and BLEU-1~\cite{papineni2002bleu}; HotpotQA and
NarrativeQA use maximum token F1 over gold answers, following common
open-domain QA practice~\cite{kwiatkowski2019natural,yang2018hotpotqa}.

\subsection{Leakage-Controlled Protocol}
The LoCoMo evaluation separates schema development from per-example scoring.
Operator-family schemas are fixed before the reported scoring pass and are
defined at the level of generic validity invariants: substrate compatibility,
answer contract, temporal/evidence boundary, count boundary, quote/reason
boundary, exact-slot restoration, and abstention. The schemas do not contain
conversation identifiers, person-specific lexical patterns, or per-example
gold-answer conditions. For each target conversation $d$, the utility ledger
and utility-threshold decision use only diagnostics from the other held-out
conversations. The gate uses the target question and retrieved state as
ordinary test-time inputs, but no gold answer, correction outcome, or aggregate
score from $d$ is used to decide whether a candidate may change an answer in
$d$.
Thus, the reported LoCoMo numbers are produced by a cross-fitted policy rather
than a policy fitted on the full evaluation split.

For the reported HotpotQA and NarrativeQA runs, the LoCoMo-defined operator
schemas, veto conditions, and threshold range are reused without
dataset-specific utility tuning or error-analysis-driven edits. These datasets
are evaluated as frozen transfer checks, with only the substrate profile
selecting which operators are eligible.
Full-split LoCoMo plots in Section~\ref{sec:results} are post-hoc summaries of
the resulting behavior; they are not used to construct per-example gates.

\subsection{Baselines}
Our primary baselines are RAG~\cite{zhu2020retrieval},
GAM~\cite{yan2025general}, and R2Mem~\cite{wang2026r}, all evaluated in the
same GPT-4o-mini API harness. RAG~\cite{zhu2020retrieval} uses a memory-free retriever-reader setting:
fixed 2048-token chunks, top-5 embedding retrieval, and one-pass answer
generation, matching the standard retriever-reader line of work~\cite{
zhu2020retrieval,karpukhin2020dense,izacard2021leveraging}. GAM and R2Mem
are included as deep-search and experience-reuse baselines.

We also include Mem0~\cite{chhikara2025mem0},
A-Mem~\cite{xu2026mem}, MemoryOS~\cite{kang2025memory},
LightMem, and Memory R1~\cite{yan2026memory}. The
structured-memory baselines span dynamic memory extraction, agentic memory
linking, hierarchical memory operating systems, lightweight memory
consolidation, and RL-trained memory operations~\cite{chhikara2025mem0,
xu2026mem,kang2025memory,yan2026memory}. We evaluate
matched-backbone reproductions under the same LoCoMo split, answerer,
embedding model, scorer, and API budget. We also discuss graph-structured and
operating-system-style memory systems as related comparators~\cite{
rasmussen2025zep,packer2023memgpt,li2025memos}.

\subsection{Implementation Details}
The main cross-dataset results use GPT-4o-mini for memory search and answer
generation, and OpenAI \texttt{text-embedding-3-small} for dense retrieval.
The Qwen2.5 LoCoMo runs use OpenAI-compatible local vLLM endpoints and local
BGE-M3 embeddings~\cite{chen2024bge}. All methods share dataset loaders,
metrics, retrieval interfaces, and scoring scripts. Hardware-dependent
components are replaced by API-friendly dense retrieval and pure-Python
BM25~\cite{robertson2009probabilistic}.

\section{Results and Analysis}
\label{sec:results}

\subsection{RQ1: Overall Performance}
\begin{table*}[!t]
\caption{LoCoMo results under GPT-4o-mini and
Qwen2.5-7B-Instruct. We report token F1 and BLEU-1 on the 1,388-question
evaluation split.}
\label{tab:main}
\centering
\scriptsize
\setlength{\tabcolsep}{3pt}
\resizebox{0.8\textwidth}{!}{%
\begin{tabular}{llrrrrrrrrrr}
\toprule
Model & Method
& \multicolumn{2}{c}{Overall}
& \multicolumn{2}{c}{Multi-hop}
& \multicolumn{2}{c}{Temporal}
& \multicolumn{2}{c}{Open}
& \multicolumn{2}{c}{Single-hop} \\
\cmidrule(lr){3-4}\cmidrule(lr){5-6}\cmidrule(lr){7-8}\cmidrule(lr){9-10}\cmidrule(lr){11-12}
& & F1 & BLEU & F1 & BLEU & F1 & BLEU & F1 & BLEU & F1 & BLEU \\
\midrule
\multirow{9}{*}{GPT-4o-mini}
& RAG~\cite{zhu2020retrieval} & 39.92 & 34.52 & 33.75 & 24.33 & 31.52 & 26.86 & 25.21 & 20.42 & 46.60 & 42.16 \\
& Mem0~\cite{chhikara2025mem0} & 48.27 & 42.10 & 30.73 & 20.91 & 52.14 & 46.13 & 31.56 & 25.70 & 54.33 & 49.25 \\
& A-Mem~\cite{xu2026mem} & 46.59 & 40.55 & 32.34 & 23.77 & 49.34 & 43.20 & 30.33 & 25.00 & 51.94 & 46.70 \\
& MemoryOS~\cite{kang2025memory} & 45.26 & 39.28 & 29.17 & 20.00 & 52.67 & 46.62 & 30.46 & 24.71 & 49.35 & 44.40 \\
& LightMem & 45.80 & 39.90 & 29.01 & 20.48 & 47.50 & 41.92 & 26.71 & 21.53 & 52.67 & 47.43 \\
& Memory-R1~\cite{yan2026memory} & 48.53 & 42.60 & 33.20 & 24.08 & 52.55 & 46.57 & 31.14 & 24.81 & 53.89 & 49.06 \\
& GAM~\cite{yan2025general} & 54.50 & 48.11 & 42.32 & 33.67 & 60.98 & 54.95 & 30.62 & 24.60 & 58.64 & 52.80 \\
& R2Mem~\cite{wang2026r} & 53.88 & 47.45 & 44.07 & 35.73 & 58.08 & 51.99 & 30.94 & 25.70 & 57.98 & 51.92 \\
& \method{} (ours) & \textbf{60.35} & \textbf{53.76} & \textbf{52.26} & \textbf{42.91} & \textbf{65.21} & \textbf{59.59} & \textbf{40.77} & \textbf{35.53} & \textbf{63.29} & \textbf{57.10} \\
\midrule
\multirow{9}{*}{Qwen2.5-7B}
& RAG~\cite{zhu2020retrieval} & 38.27 & 32.87 & 30.70 & 21.62 & 31.38 & 26.84 & 17.73 & 14.64 & 45.48 & 40.70 \\
& Mem0~\cite{chhikara2025mem0} & 37.68 & 32.16 & 28.48 & 20.41 & \textbf{39.45} & \textbf{33.45} & 23.15 & 18.89 & 41.58 & 36.92 \\
& A-Mem~\cite{xu2026mem} & 34.50 & 28.49 & 30.39 & 21.52 & 37.84 & 32.69 & 20.50 & 17.26 & 36.10 & 30.41 \\
& MemoryOS~\cite{kang2025memory} & 34.42 & 29.22 & 27.46 & 19.78 & 38.04 & 32.53 & 20.92 & 17.22 & 36.80 & 32.36 \\
& LightMem & 36.09 & 30.84 & 29.25 & 21.38 & 36.28 & 31.14 & 18.30 & 14.85 & 40.15 & 35.51 \\
& Memory-R1~\cite{yan2026memory} & 37.21 & 31.60 & 29.68 & 21.20 & 38.41 & 32.76 & 20.70 & 16.52 & 40.98 & 36.16 \\
& GAM~\cite{yan2025general} & 44.76 & 38.56 & \textbf{37.29} & \textbf{28.66} & 27.79 & 22.90 & \textbf{26.86} & \textbf{23.18} & 55.36 & 49.19 \\
& R2Mem~\cite{wang2026r} & 43.15 & 37.15 & 35.62 & 27.66 & 27.68 & 23.40 & 21.05 & 16.93 & 53.68 & 47.47 \\
& \method{} (ours) & \textbf{46.63} & \textbf{40.15} & 36.80 & 28.02 & 34.69 & 30.08 & 24.18 & 20.59 & \textbf{56.63} & \textbf{49.90} \\
\bottomrule
\end{tabular}
}
\end{table*}

Table~\ref{tab:main} shows that \method{} is the strongest overall LoCoMo
method under both completed backbones. With GPT-4o-mini, it improves over the
strongest compared baseline by +5.85 F1 and +5.65 BLEU-1. With Qwen2.5-7B,
the corresponding gains are +1.87 F1 and +1.59 BLEU-1. The category breakdown
shows the largest improvements on temporal, single-hop, and GPT-4o-mini
open-domain questions, which are the categories most affected by answer
contract and evidence-boundary errors.

\begin{table}[t]
\caption{Cross-dataset long-context QA check under GPT-4o-mini. NarrativeQA
uses the 300-question average-87K subset; HotpotQA reports the 56K, 224K, and
448K context regimes.}
\label{tab:cross}
\centering
\scriptsize
\begin{tabular}{lrrrr}
\toprule
Method & NQA 87K & HP 56K & HP 224K & HP 448K \\
\midrule
RAG~\cite{zhu2020retrieval} & 32.54 & 46.86 & 31.52 & 25.98 \\
GAM~\cite{yan2025general} & 38.66 & 64.07 & 60.19 & 59.36 \\
R2Mem~\cite{wang2026r} & 37.24 & 62.74 & 61.27 & 61.70 \\
\method{} (ours) & \textbf{39.37} & \textbf{66.24} & \textbf{61.58} & \textbf{61.87} \\
\bottomrule
\end{tabular}
\end{table}

Table~\ref{tab:cross} extends RQ1 beyond LoCoMo with a secondary check on the
two non-LoCoMo datasets. The same GPT-4o-mini backbone is used throughout.
\method{} achieves the best score in NarrativeQA and all three HotpotQA context
regimes under the frozen schema and threshold setting described above. In the
448K HotpotQA setting, the gate preserves the strongest base branch when
intervention conditions are not satisfied and applies bounded answer-contract
edits only when they pass the checks.

\subsection{RQ2: Scaling Behavior}
Under Qwen2.5-3B, RAG, GAM, R2Mem, and \method{} obtain 17.45, 30.39,
32.88, and 34.65 overall F1, respectively, so \method{} gains 1.76 F1 over
the strongest baseline. Table~\ref{tab:main} shows a corresponding 1.87-F1
gain under Qwen2.5-7B. The same gate and operator schemas are used across
backbones.

\subsection{Intervention Frequency}
The validity gate often abstains. On HotpotQA, the cross-dataset overlay changes
only a small number of answers: zero at 56K, 11 at 224K, and five at 448K after
the long-context boundary selects the most reliable base branch. On NarrativeQA,
only 16 of 300 answers are changed, yet F1 increases from 38.73 to 39.37. This
shows that the gains do not require broad rewriting. On LoCoMo, answer-contract
validation changes fewer than 10\% of examples in each category, temporal
validation changes 2.1\% of temporal questions, and open-inference validation
changes 14.5\% of open-domain questions.

\subsection{Failure Patterns}
The strongest LoCoMo gains come from correcting answer-contract mismatches:
off-slot attributes, lost temporal relations, dropped reason or quote
components, over-specific durations, and unsupported broad count/date
overrides. Broad GPT-based candidate auditing and broad raw-localizer overlays
are included only as diagnostic negative controls and are never eligible to
alter answers in the evaluated policy. The full-split summaries characterize
the cross-fitted policy after evaluation; they are not used to decide whether
any target-conversation answer should be changed.

\begin{figure*}[!t]
\centering
\includegraphics[width=0.90\textwidth]{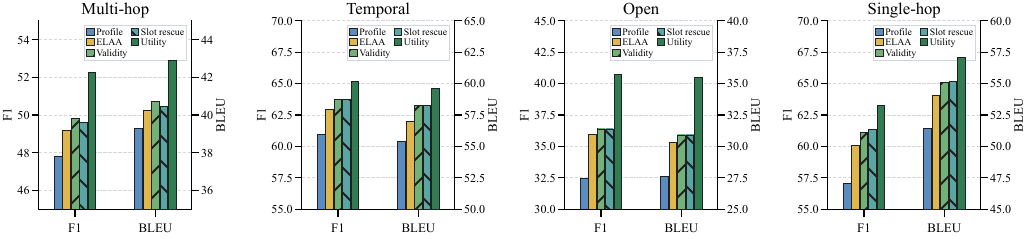}
\caption{Category-level component ablation on LoCoMo. Bars correspond
to five diagnostic checkpoints: profile, ELAA arbitration, validity-boundary
checking, exact-slot rescue, and utility-aware selection. Conceptually, ELAA
arbitration and exact-slot rescue are two candidate-intervention generators
within the same high-level component.}
\label{fig:ablation}
\end{figure*}

\begin{figure*}[!t]
\centering
\includegraphics[width=0.84\textwidth]{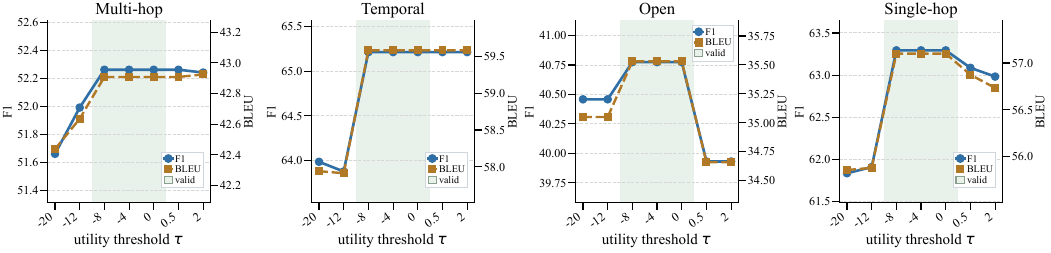}
\caption{Sensitivity analysis for utility-gated intervention on LoCoMo. Each
point uses conversation-level cross-fitting; the shaded region marks the
pre-specified operating range rather than a per-conversation optimum.}
\label{fig:utility-threshold}
\end{figure*}

\begin{figure*}[!t]
\centering
\includegraphics[width=0.84\textwidth]{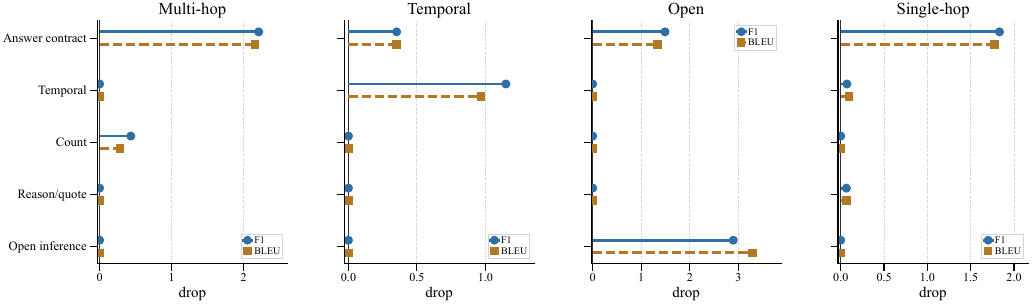}
\caption{Leave-one-boundary-family ablation on LoCoMo. Each panel reports the
score drop from full \method{} after disabling one validation family.}
\label{fig:boundary-family}
\end{figure*}

\subsection{RQ3: Ablation Study}
Fig.~\ref{fig:ablation} summarizes the component trajectory. Answer arbitration
gives the largest early jump because many LoCoMo errors are answer-contract
violations rather than retrieval failures. Validity-boundary checks then
suppress candidate edits whose preconditions do not hold. Exact slot rescue
fixes cases where raw dialogue contains the answer-bearing phrase but summary
generation drops it, while the final utility overlay accepts only candidate
families with positive empirical effect. Candidate generation thus increases
recall, boundary checks filter applicability errors, and utility selection
decides whether an intervention is worth executing; the mostly monotonic
category trajectories support this ordering. Threshold sweeps further show a
stable plateau around the selected operating range: permissive thresholds
admit high-variance edits, whereas overly strict thresholds suppress useful
low-frequency corrections. Leave-one-family diagnostics agree with the design:
answer-contract checks dominate multi- and single-hop gains, temporal checks
matter most for temporal questions, and open-inference checks matter most for
open-domain questions (Figs.~\ref{fig:utility-threshold}
and~\ref{fig:boundary-family}).

\subsection{Why Flat Experience Can Regress}
Flat experience reuse can fail even when the retrieved guidance is
semantically relevant. A ``prefer concise answer'' operator can help direct
location questions but hurt reason or feeling questions; a ``latest event''
operator can help current-state questions but hurt date-constrained questions;
and counting operators can help exact counts but hurt duration or frequency
questions. Relevance to the task type therefore does not guarantee
compatibility with the current answer contract, context regime, or evidence
state.

\subsection{Qualitative Failure Taxonomy}
We group corrected LoCoMo errors into six recurring classes: off-slot
substitution, where the answer gives the wrong attribute; temporal drift, where
the model copies a plausible but wrongly constrained date or current state;
count looseness, where exact-count questions receive vague frequencies; quote
and feeling loss, where paraphrase drops the answer-bearing expression;
open-domain entity confusion, where the model must infer an external entity
from dialogue evidence; and over-compression, where a multi-clause answer is
shortened until it drops a necessary condition. Corrected failures concentrate
around answer-slot and granularity errors, followed by temporal, quote/feeling,
and open-inference errors, matching the gate's emphasis on post-retrieval
validity rather than indiscriminate answer rewriting.

\section{Conclusion}

We introduced \method{}, a training-free gate for deciding when retrieved
experience should be allowed to change a memory-search answer. The method
represents experience as typed operators and checks substrate compatibility,
answer contract, evidence boundary, and cross-fitted utility before applying an
operator. Across LoCoMo, HotpotQA, and NarrativeQA, this selective policy
improves over retrieval, deep-search, and experience-reuse baselines. The main
practical takeaway is simple: reusable memory should include the conditions for
use, not only the instruction to reuse.

\bibliographystyle{IEEEtran}
\bibliography{references}

@article{wang2026r,
  title={R\^{} 2-Mem: Reflective Experience for Memory Search},
  author={Wang, Xinyuan and Mao, Wenyu and Wu, Junkang and Wang, Xiang and He, Xiangnan},
  journal={arXiv preprint arXiv:2605.13486},
  year={2026}
}

@article{yan2025general,
  title={General agentic memory via deep research},
  author={Yan, BY and Li, Chaofan and Qian, Hongjin and Lu, Shuqi and Liu, Zheng},
  journal={arXiv preprint arXiv:2511.18423},
  year={2025}
}

@article{zhu2020retrieval,
  title={Retrieval-augmented generation for knowledge-intensive nlp tasks},
  author={Zhu, Zengliang and Zhang, Yuchen and Peng, Jordan},
  year={2020}
}

@inproceedings{guu2020retrieval,
  title={Retrieval augmented language model pre-training},
  author={Guu, Kelvin and Lee, Kenton and Tung, Zora and Pasupat, Panupong and Chang, Mingwei},
  booktitle={International conference on machine learning},
  pages={3929--3938},
  year={2020},
  organization={PMLR}
}

@inproceedings{karpukhin2020dense,
  title={Dense passage retrieval for open-domain question answering},
  author={Karpukhin, Vladimir and Oguz, Barlas and Min, Sewon and Lewis, Patrick and Wu, Ledell and Edunov, Sergey and Chen, Danqi and Yih, Wen-tau},
  booktitle={Proceedings of the 2020 conference on empirical methods in natural language processing (EMNLP)},
  pages={6769--6781},
  year={2020}
}

@inproceedings{izacard2021leveraging,
  title={Leveraging passage retrieval with generative models for open domain question answering},
  author={Izacard, Gautier and Grave, Edouard},
  booktitle={Proceedings of the 16th conference of the european chapter of the association for computational linguistics: main volume},
  pages={874--880},
  year={2021}
}

@book{robertson2009probabilistic,
  title={The probabilistic relevance framework: BM25 and beyond},
  author={Robertson, Stephen and Zaragoza, Hugo},
  volume={4},
  year={2009},
  publisher={Now Publishers Inc}
}

@inproceedings{reimers2019sentence,
  title={Sentence-bert: Sentence embeddings using siamese bert-networks},
  author={Reimers, Nils and Gurevych, Iryna},
  booktitle={Proceedings of the 2019 conference on empirical methods in natural language processing and the 9th international joint conference on natural language processing (EMNLP-IJCNLP)},
  pages={3982--3992},
  year={2019}
}

@article{chen2024bge,
  title={Bge m3-embedding: Multi-lingual, multi-functionality, multi-granularity text embeddings through self-knowledge distillation},
  author={Chen, Jianlv and Xiao, Shitao and Zhang, Peitian and Luo, Kun and Lian, Defu and Liu, Zheng},
  journal={arXiv preprint arXiv:2402.03216},
  volume={4},
  number={5},
  year={2024}
}

@article{shinn2024reflexion,
  title={Reflexion: Language agents with verbal reinforcement learning, 2023},
  author={Shinn, Noah and Cassano, Federico and Berman, Edward and Gopinath, Ashwin and Narasimhan, Karthik and Yao, Shunyu},
  journal={URL https://arxiv. org/abs/2303.11366},
  volume={8},
  year={2024}
}

@article{madaan2023self,
  title={Self-refine: Iterative refinement with self-feedback},
  author={Madaan, Aman and Tandon, Niket and Gupta, Prakhar and Hallinan, Skyler and Gao, Luyu and Wiegreffe, Sarah and Alon, Uri and Dziri, Nouha and Prabhumoye, Shrimai and Yang, Yiming and others},
  journal={Advances in neural information processing systems},
  volume={36},
  pages={46534--46594},
  year={2023}
}

@misc{wang2023voyageropenendedembodiedagent,
      title={Voyager: An Open-Ended Embodied Agent with Large Language Models}, 
      author={Guanzhi Wang and Yuqi Xie and Yunfan Jiang and Ajay Mandlekar and Chaowei Xiao and Yuke Zhu and Linxi Fan and Anima Anandkumar},
      year={2023},
      eprint={2305.16291},
      archivePrefix={arXiv},
      primaryClass={cs.AI},
      url={https://arxiv.org/abs/2305.16291}, 
}

@inproceedings{park2023generative,
  title={Generative agents: Interactive simulacra of human behavior},
  author={Park, Joon Sung and O'Brien, Joseph and Cai, Carrie Jun and Morris, Meredith Ringel and Liang, Percy and Bernstein, Michael S},
  booktitle={Proceedings of the 36th annual acm symposium on user interface software and technology},
  pages={1--22},
  year={2023}
}

@article{packer2023memgpt,
  title={MemGPT: towards LLMs as operating systems.},
  author={Packer, Charles and Fang, Vivian and Patil, Shishir\_G and Lin, Kevin and Wooders, Sarah and Gonzalez, Joseph\_E},
  year={2023},
  publisher={ArXiv}
}

@article{chhikara2025mem0,
  title={Mem0: Building production-ready ai agents with scalable long-term memory},
  author={Chhikara, Prateek and Khant, Dev and Aryan, Saket and Singh, Taranjeet and Yadav, Deshraj},
  journal={arXiv preprint arXiv:2504.19413},
  year={2025}
}

@article{xu2026mem,
  title={A-mem: Agentic memory for llm agents},
  author={Xu, Wujiang and Liang, Zujie and Mei, Kai and Gao, Hang and Tan, Juntao and Zhang, Yongfeng},
  journal={Advances in Neural Information Processing Systems},
  volume={38},
  pages={17577--17604},
  year={2026}
}

@inproceedings{kang2025memory,
  title={Memory os of ai agent},
  author={Kang, Jiazheng and Ji, Mingming and Zhao, Zhe and Bai, Ting},
  booktitle={Proceedings of the 2025 Conference on Empirical Methods in Natural Language Processing},
  pages={25972--25981},
  year={2025}
}

@inproceedings{yan2026memory,
  title={Memory-r1: Enhancing large language model agents to manage and utilize memories via reinforcement learning},
  author={Yan, Sikuan and Yang, Xiufeng and Huang, Zuchao and Nie, Ercong and Ding, Zifeng and Li, Zonggen and Ma, Xiaowen and Bi, Jinhe and Kersting, Kristian and Pan, Jeff Z and others},
  booktitle={Proceedings of the 64th Annual Meeting of the Association for Computational Linguistics (Volume 1: Long Papers)},
  pages={12805--12825},
  year={2026}
}

@article{rasmussen2025zep,
  title={Zep: a temporal knowledge graph architecture for agent memory},
  author={Rasmussen, Preston and Paliychuk, Pavlo and Beauvais, Travis and Ryan, Jack and Chalef, Daniel},
  journal={arXiv preprint arXiv:2501.13956},
  year={2025}
}

@article{li2025memos,
  title={Memos: An operating system for memory-augmented generation (mag) in large language models},
  author={Li, Zhiyu and Song, Shichao and Wang, Hanyu and Niu, Simin and Chen, Ding and Yang, Jiawei and Xi, Chenyang and Lai, Huayi and Zhao, Jihao and Wang, Yezhaohui and others},
  journal={arXiv preprint arXiv:2505.22101},
  year={2025}
}

@inproceedings{banerjee2026apex,
  title={APEX-MEM: Agentic Semi-Structured Memory with Temporal Reasoning for Long-Term Conversational AI},
  author={Banerjee, Pratyay and Moshtaghi, Masud and Subramanian, Shivashankar and Misra, Amita and Chadha, Ankit},
  booktitle={Proceedings of the 64th Annual Meeting of the Association for Computational Linguistics (Volume 1: Long Papers)},
  pages={16470--16489},
  year={2026}
}

@inproceedings{yang2018hotpotqa,
  title={HotpotQA: A dataset for diverse, explainable multi-hop question answering},
  author={Yang, Zhilin and Qi, Peng and Zhang, Saizheng and Bengio, Yoshua and Cohen, William and Salakhutdinov, Ruslan and Manning, Christopher D},
  booktitle={Proceedings of the 2018 conference on empirical methods in natural language processing},
  pages={2369--2380},
  year={2018}
}

@article{kovcisky2018narrativeqa,
  title={The narrativeqa reading comprehension challenge},
  author={Ko{\v{c}}isk{\`y}, Tom{\'a}{\v{s}} and Schwarz, Jonathan and Blunsom, Phil and Dyer, Chris and Hermann, Karl Moritz and Melis, G{\'a}bor and Grefenstette, Edward},
  journal={Transactions of the Association for Computational Linguistics},
  volume={6},
  pages={317--328},
  year={2018},
  publisher={MIT Press One Rogers Street, Cambridge, MA 02142-1209, USA journals-info~…}
}

@article{liu2023lost,
  title={Lost in the middle: How language models use long contexts},
  author={Liu, Nelson F and Lin, Kevin and Hewitt, John and Paranjape, Ashwin and Bevilacqua, Michele and Petroni, Fabio and Liang, Percy},
  journal={arXiv preprint arXiv:2307.03172},
  year={2023}
}

@article{bai2023longbench,
  title={Longbench: A bilingual, multitask benchmark for long context understanding},
  author={Bai, Yushi and Lv, Xin and Zhang, Jiajie and Lyu, Hongchang and Tang, Jiankai and Huang, Zhidian and Du, Zhengxiao and Liu, Xiao and Zeng, Aohan and Hou, Lei and others},
  journal={arXiv preprint arXiv:2308.14508},
  year={2023}
}

@inproceedings{yen2025helmet,
  title={HELMET: How to evaluate long-context models effectively and thoroughly},
  author={Yen, Howard and Gao, Tianyu and Hou, Minmin and Ding, Ke and Fleischer, Daniel and Izsak, Peter and Wasserblat, Moshe and Chen, Danqi},
  booktitle={The Thirteenth International Conference on Learning Representations},
  year={2025}
}

@article{wu2024longmemeval,
  title={Longmemeval: Benchmarking chat assistants on long-term interactive memory},
  author={Wu, Di and Wang, Hongwei and Yu, Wenhao and Zhang, Yuwei and Chang, Kai-Wei and Yu, Dong},
  journal={arXiv preprint arXiv:2410.10813},
  year={2024}
}

@inproceedings{li2026locomo,
  title={Locomo-plus: Beyond-factual cognitive memory evaluation framework for llm agents},
  author={Li, Yifei and Guo, Weidong and Zhang, Lingling and Xu, Rongman and Huang, Muye and Liu, Hui and Xu, Lijiao and Xu, Yu and Liu, Jun},
  booktitle={Proceedings of the 64th Annual Meeting of the Association for Computational Linguistics (Volume 1: Long Papers)},
  pages={25085--25100},
  year={2026}
}

@article{kwiatkowski2019natural,
  title={Natural questions: a benchmark for question answering research},
  author={Kwiatkowski, Tom and Palomaki, Jennimaria and Redfield, Olivia and Collins, Michael and Parikh, Ankur and Alberti, Chris and Epstein, Danielle and Polosukhin, Illia and Devlin, Jacob and Lee, Kenton and others},
  journal={Transactions of the Association for Computational Linguistics},
  volume={7},
  pages={453--466},
  year={2019},
  publisher={MIT Press One Rogers Street, Cambridge, MA 02142-1209, USA journals-info~…}
}

@inproceedings{mallen2023not,
  title={When not to trust language models: Investigating effectiveness of parametric and non-parametric memories},
  author={Mallen, Alex and Asai, Akari and Zhong, Victor and Das, Rajarshi and Khashabi, Daniel and Hajishirzi, Hannaneh},
  booktitle={Proceedings of the 61st annual meeting of the association for computational linguistics (volume 1: Long papers)},
  pages={9802--9822},
  year={2023}
}

@article{graves2012long,
  title={Long short-term memory},
  author={Graves, Alex},
  journal={Supervised sequence labelling with recurrent neural networks},
  pages={37--45},
  year={2012},
  publisher={Springer}
}

@article{graves2014neural,
  title={Neural turing machines},
  author={Graves, Alex and Wayne, Greg and Danihelka, Ivo},
  journal={arXiv preprint arXiv:1410.5401},
  year={2014}
}

@article{weston2014memory,
  title={Memory networks},
  author={Weston, Jason and Chopra, Sumit and Bordes, Antoine},
  journal={arXiv preprint arXiv:1410.3916},
  year={2014}
}

@article{sukhbaatar2015end,
  title={End-to-end memory networks},
  author={Sukhbaatar, Sainbayar and Szlam, Arthur and Weston, Jason and Fergus, Rob},
  journal={arXiv preprint arXiv:1503.08895},
  year={2015}
}

@inproceedings{papineni2002bleu,
  title={Bleu: a method for automatic evaluation of machine translation},
  author={Papineni, Kishore and Roukos, Salim and Ward, Todd and Zhu, Wei-Jing},
  booktitle={Proceedings of the 40th annual meeting of the Association for Computational Linguistics},
  pages={311--318},
  year={2002}
}

\end{document}